\documentclass[sigconf]{acmart}

\AtBeginDocument{%
  }

\copyrightyear{2023}
\acmYear{2023}
\setcopyright{acmlicensed}
\acmConference[ARTMAN '23]{Proceedings of the 2023 Workshop on Recent Advances in Resilient and Trustworthy ML Systems in Autonomous Networks}{November 30, 2023}{Copenhagen, Denmark}
\acmBooktitle{Proceedings of the 2023 Workshop on Recent Advances in Resilient and Trustworthy ML Systems in Autonomous Networks (ARTMAN '23), November 30, 2023, Copenhagen, Denmark}
\acmPrice{15.00} 
\acmISBN{979-8-4007-0265-5/23/11}
\acmDOI{10.1145/3605772.3624002}

\usepackage{xcolor}
\usepackage[export]{adjustbox}
\usepackage{balance}
\usepackage{graphicx}
\usepackage{array}%
\usepackage{algorithm} 
\usepackage{algpseudocode}
\usepackage{multirow}
\usepackage{lscape}
\usepackage{bbm}
\usepackage{stmaryrd}
\usepackage[utf8]{inputenc}
\usepackage{fancyhdr}
\usepackage{soul}
\usepackage{microtype}

\usepackage{amsmath}
\DeclareMathOperator*{\argmax}{argmax}
\DeclareMathOperator*{\argmin}{argmin}

\usepackage{pifont}
\colorlet{linecol}{black!75}

\usepackage{forest}
\usepackage{newfloat}
    \DeclareFloatingEnvironment[fileext=lod]{Figure} %for adding captions

    \forestset{/.style=
               {
                 {for tree=
                   {parent anchor=south, 
                    child anchor=north,
                    align=center
                    l+=1.5cm}
                  }
               }
              }

\begin{document}

%\title{Breaking Boundaries: Balanced Performance and Robustness of Deep Forecasting Models for Wireless Traffic Prediction}
\title{Breaking Boundaries: Balancing Performance and Robustness in Deep Wireless Traffic Forecasting}

\author{Romain Ilbert}
\orcid{}
\affiliation{%
  \institution{University Paris-Cité \& Huawei Paris Research Center}
  \streetaddress{}
  \city{Paris}
  \country{France}
  \postcode{}
}
\email{romain.ilbert@hotmail.fr}

\author{Thai V. Hoang}
\orcid{}
\affiliation{%
  \institution{Huawei Paris Research Center}
  \streetaddress{}
  \city{Paris}
  \country{France}
  \postcode{}
}
\email{thai.v.hoang@huawei.com}

\author{Zonghua Zhang}
\orcid{}
\affiliation{%
  \institution{Huawei Paris Research Center}
  \streetaddress{}
  \city{Paris}
  \country{France}
  \postcode{}
}
\email{zonghua.zhang@huawei.com}

\author{Themis Palpanas}
\orcid{}
\affiliation{%
  \institution{University of Paris}
  \streetaddress{}
  \city{Paris}
  \country{France}
  \postcode{}
}
\email{themis@mi.parisdescates.fr}
%%
%% By default, the full list of authors will be used in the page
%% headers. Often, this list is too long, and will overlap
%% other information printed in the page headers. This command allows
%% the author to define a more concise list
%% of authors' names for this purpose.
% \renewcommand{\shortauthors}{Ilbert et al.}
\renewcommand{\shortauthors}{Romain Ilbert, Thai V. Hoang, Zonghua Zhang, \& Themis Palpanas} 
%%
%% The abstract is a short summary of the work to be presented in the
%% article.

\begin{abstract}
Balancing the trade-off between accuracy and robustness is a long-standing challenge in time series forecasting.
While most of existing robust algorithms have achieved certain suboptimal performance on clean data, sustaining the same performance level in the presence of data perturbations remains extremely hard. 
In this paper, we study a wide array of perturbation scenarios and propose novel defense mechanisms against adversarial attacks using real-world telecom data.
We compare our strategy against two existing adversarial training algorithms under a range of maximal allowed perturbations, defined using $\ell_{\infty}$-norm, $\in [0.1,0.4]$.
Our findings reveal that our hybrid strategy, which is composed of a classifier to detect adversarial examples, a denoiser to eliminate noise from the perturbed data samples, and a standard forecaster, achieves the best performance on both clean and perturbed data.
Our optimal model can retain up to $92.02\%$ the performance of the original forecasting model in terms of Mean Squared Error (MSE) on clean data, while being more robust than the standard adversarially trained models on perturbed data.
Its MSE is 2.71$\times$ and 2.51$\times$ lower than those of comparing methods on normal and perturbed data, respectively. 
In addition, the components of our models can be trained in parallel, resulting in better computational efficiency.
Our results indicate that we can optimally balance the trade-off between the performance and robustness of forecasting models by improving the classifier and denoiser, even in the presence of sophisticated and destructive poisoning attacks.
\end{abstract}

%%
%% The code below is generated by the tool at http://dl.acm.org/ccs.cfm.
%% Please copy and paste the code instead of the example below.
%%


%%
%% Keywords. The author(s) should pick words that accurately describe
%% the work being presented. Separate the keywords with commas.
%% A "teaser" image appears between the author and affiliation
%% information and the body of the document, and typically spans the
%% page.

% \received{21 July 2023}
% \received[Accepted]{28 August 2023}

%%
%% This command processes the author and affiliation and title
%% information and builds the first part of the formatted document.
\maketitle

\section{Introduction}
\label{sec:introduction}

Time series forecasting  has been widely applied in various domains, such as finance, economics, healthcare, climate change, energy management, and telecommunications \cite{Hallac2017, choi2016doctor, MUDELSEE2019310, global_energy_compet, zhang2019deep}.
For example, wireless traffic forecasting has found its promising role in resource allocation, traffic engineering, and network security \cite{resource_alloc_load_pred, resource_alloc_vehicular, resource_allocation, traffic_usage_forecasting, wireless_traffic_clustering, arjoune_artificial_2020}. 
%It entails forecasting wireless network traffic using historical data to enable informed decision-%making and performance optimization.
In particular, one can obtain better load balancing and capacity planning by predicting future network traffic demand using historical data. 
Proactive maintenance and fault management can also be enabled by predicting network failures~\cite{botta2016integration}.
Recently, deep learning has been developed for time series forecasting, showing significant advantages over the classical methods in terms of performance, robustness, and generalizability \cite{rangapuram2018deep, salinas2020deepar, lim2020temporal, wang2019deep}. 
However, as those deep learning models applied in many other domains (e.g., Computer Vision, Natural Language Processing), whose performance and usability could be seriously undermined in the face of data poisoning attacks \cite{szegedy2014intriguing, biggio2013poisoning, goodfellow2015explaining, papernot2015limitations, chen2017targeted}, the ones applied for time series forecasting are not immune from those attacks. 
One of the fundamental differences to be carefully considered is that time series data always exhibits temporal dependencies, thereby requiring special modeling techniques and attack strategies \cite{shumway2017time, hamilton1994time, hyndman2018forecasting, box1976time}.

Specifically, data poisoning refers to the attack that is intended to deliberately manipulate the data fed to a machine learning model \cite{biggio2013poisoning}, eventually undermining its performance. 
Data poisoning attack can be further classified into label flipping, data injection, and data modification \cite{xiao2015feature}, each of which has its own assumptions and requirements about the attacker's capabilities and goals \cite{chen2017zoo}. 
For example, an attacker may have full or partial knowledge of the model architecture, access to the training data, or the ability to inject malicious samples into the dataset.
Depending on the forecasting models and application scenarios, the consequence of a poisoning attack can be extremely destructive. 
For instance, a successful poisoning attack to a financial forecasting model could lead to financial losses and market instability.
In more critical scenarios like healthcare, where time series forecasting models are used for patient monitoring, disease diagnosis, and treatment planning, poisoning attacks can cause incorrect diagnoses, delayed treatments, and even life-threatening situations \cite{selvakkumar2021addressing, liu2017trojaning}.
Similarly, poisoned wireless traffic forecasting models could result in deteriorated network performance and even network failures  \cite{mei2015using}. 
As a matter of fact, all these attacks have been demonstrated to be effective in various domains, such as Computer Vision (CV), Natural Language Processing (NLP) \cite{biggio2014security, papernot2016practical, karim2019adversarial}. 
But their feasibility and effectiveness in time series forecasting have been studied much less than they deserve \cite{liu2022practical, liu2023robust, zheng_poisoning_2022}. 
In addition, the thread models in most of the existing works are oversimplified, while the concern is limited to the trade-off between model's performance and robustness.  
Taking wireless traffic prediction as a specific scenario, we make the following contributions in this paper:
\begin{itemize}
    \item A comprehensive examination of data poisoning attacks on deep learning-based time series forecasting models.
    \item A novel defense mechanism against these attacks that involves one classifier to identify perturbed data and one denoiser to remove perturbations from those data.
    \item A new bi-level masking attack strategy under extreme adversarial conditions, with $\ell_{\infty}$-norm $\to 0.4$ across all sequences and their individual steps.
    \item  Our optimal model preserves up to 92.02\% of the original forecasting model's MSE on clean data. Its MSE is up to $2.71\times$ and $2.51\times$ lower than those of existing methods on clean and perturbed data, respectively.
\end{itemize}
   
The effectiveness of our proposed defense mechanism has been validated on real-world telecom dataset. Experimental results show that it can significantly improve the performance of wireless traffic prediction models, even under strong poisoning attacks, while maintaining its performance on clean data.

The rest of this paper is organized as follows. 
Section II reviews related work on wireless traffic prediction and adversarial attacks. 
Section III and IV presents in detail the proposed attack and defense mechanisms, respectively. 
Section V describes the experimental setup and presents the evaluation results. 
Finally, Section VI concludes the paper with discussions and future research directions.

\section{Related work}
\label{sec:background}

We review in this section existing works that are closely related to ours.
This includes time series prediction, data poisoning attacks and the corresponding countermeasures.

\subsection{Wireless Traffic Prediction}
\label{subsec:wireless}

Wireless traffic prediction has been extensively studied due to its importance in various network applications. 
So far, various techniques have been proposed to model and forecast wireless traffic effectively.
Traditional forecasting methods for time series, such as ARIMA models, exponential smoothing state-space models, and STL, are frequently utilized to model linear relationships, trends, and seasonality in data \cite{box1976time, hyndman2008forecasting, cleveland1990stl, chatfield2003analysis}. 
Despite their widespread use, these methods exhibit certain limitations. 
In particular, they may falter when dealing with highly nonlinear or intricate patterns and often necessitate manual parameter tuning, a process that can be laborious and may not guarantee optimal performance.

To address these limitations, deep learning models have emerged as popular choices for time series forecasting tasks. 
Recurrent Neural Networks (RNN), Long Short-Term Memory networks (LSTM), and Gated Recurrent Units (GRU) have been particularly successful in capturing complex temporal dependencies and nonlinear relationships in time series data \cite{hochreiter1997long,cho2014learning}. 
These models can automatically learn intricate patterns, often resulting in improved predictive performance compared to traditional methods.
More recently, Convolutional Neural Networks (CNN) showed promising results in capturing local and global patterns in time series data \cite{borovykh2017conditional, bai2018empirical}. 
Transformer-based architectures, which were initially designed for NLP tasks, have also been adapted for time series forecasting \cite{vaswani2017attention}.
Considering the application scenario and characteristics of the used dataset, we choose LSTM for the implementation of the baseline forecaster in this work.

\subsection{Data Poisoning Attacks}
\label{subsec:poisoning}

To date, several types of attacks have been identified in the literature.
For example, an attacker can alter the labels (i.e., label flipping) of a subset of training data \cite{xiao2015feature}, or introduce malicious samples (i.e., data injection) into the training dataset \cite{steinhardt2017certified}, and even subtly alter existing data points (i.e., data modification) to mislead the forecasting task during training \cite{biggio2013poisoning}. 
In order to achieve these attacks, an adversary should possess particular capabilities and goals, such as the knowledge about model's architecture, and the access privileges to training data \cite{chen2017zoo}.
For example, Projected Gradient Descent (PGD) aims at identifying the most effective adversarial examples by maximizing the model's loss function through iteratively perturbing input samples \cite{madry2017towards, kurakin2016adversarial}.   
To successfully perform PGD, an adversary needs access to the model's architecture, the ability to compute the loss function and its gradient with respect to the input data.
This type of attack is considered as strong adversary, as it can craft adversarial examples that are more likely to fool the model than single-step methods such as Fast Gradient Sign Method (FGSM) \cite{goodfellow2015explaining}.

\subsection{Countermeasures against Poisoning Attacks}
\label{subsec:defenses}

Defending against data poisoning attacks remains a challenging and active area of research in recent years. 
To date, various defense strategies have been proposed to mitigate the effects of data poisoning attacks, and they can be classified into two categories.
The first relies on data sanitization by filtering out perturbed samples from the training data.
For example, in \cite{rubinstein2009antidote, li2015data, barreno2006can}, the authors illustrated that outlier detection techniques can be used to identify and remove potentially poisoned samples. 
However, a recent study has provided a theoretical framework and pointed out that detecting adversarial examples (i.e., poisoned data) is nearly as hard as classifying them \cite{detecting-adversarial-examples-pmlr2022}. 

Other than data sanitization, we have also seen some attempts to learning robust machine learning models, which are expected to be less sensitive to the presence of poisoned data than their normal counterparts \cite{papernot2015distillation, duchi2012randomized, steinhardt2017certified}. 

In particular, adversarial training has recently shown significant promise and gained popularity for improving the robustness of machine learning models. 
Its key idea is to train models on adversarially perturbed examples in order to force it to generalize better under adversarial conditions \cite{madry2017towards}. 
A number of strategies have been developed in~\cite{mosbach2018logit, xie2019feature, dong2018boosting} following this idea. 
For example, in \cite{zhang2019theoretically}, an approach termed TRADES was developed to balance the trade-off between the standard and robust accuracy.
It is also worth mentioning that a recent study investigated the limits of adversarial training against norm-bounded adversarial examples and provided insights into the potential trade-offs between standard and robust generalization \cite{gowal2020uncovering}.

As mentioned previously in Section \ref{subsec:poisoning}, despite tremendous efforts paid to the development of countermeasures against poisoning attacks in CV and NLP, much less work has focused on time series analytics, in particular scenarios such as speech recognition \cite{jia2017adversarial} and industrial control systems \cite{carlini2017magnet}.
Our work focuses on learning robust forecasting models for the wireless network traffic prediction problem. 
It can be considered as a new contribution to the aforementioned second category.
Unlike \cite{detecting-adversarial-examples-pmlr2022}, we demonstrate that guaranteeing robustness against adversarial attacks is as hard as classifying and denoising the adversarial samples.

\subsection{Comparison to Previous Works}
\label{subsec:comparison}

In order to distinguish our contributions from existing works, we enumerate below the four models that have been implemented for performance benchmarking in this work. 
Details of these models are presented in Fig.~\ref{fig:evaluation_frameworks}.
\begin{itemize}
    \item $M_1$: a forecaster trained with normal data;
    \item $M_2$: a forecaster trained with normal and poisoned data;
    \item $M_3$: a hybrid model composed of a classifier to identify and a denoiser to remove perturbation from poisoned data ahead of a forecaster;
    \item $M_4$: a hybrid model composed of a classifier to identify and direct normal and poisoned data to two separated forecasters.
\end{itemize}
The forecasters employed in these models use LSTM networks and share the same architecture.
It can be seen that $M_1$ serves as a baseline forecasting model.
$M_2$ embodies the approach of Madry et al.'s~\cite{madry2017towards}, which is usually used as the adversarially trained baseline forecasting model.
To the best of our knowledge, no existing work has ever incorporated a classifier and/or a denoiser into an end-to-end adversarially training framework as our proposed  $M_3$ and $M_4$ models.
In particular, the authors of \cite{zheng_poisoning_2022} employ a data sanitization approach and a statistical-based anomaly detection method to remove potential outliers.

\section{Threat Modeling}
\label{sec:attack}

Poisoning attacks in wireless traffic prediction have been formulated as machine learning problems in both centralized and distributed scenarios \cite{zheng_poisoning_2022}. 
In this work, we focus on centralized settings, due to their wide-spread implementation and application in real-world scenarios. 
In such a setting, the traffic data of different clients are transferred to a common server in order to train a global forecasting model using a standard model optimization process. 

Let's consider $F_1$, a forecaster trained on normal data using a standard empirical risk minimization (ERM) scheme. 
It is defined as $f_1(x;\theta)$, where $x$ is the input data and $\theta$ represents $F_1$'s parameters. 
The loss function $\mathcal{L}$ used in the training is defined as the MSE between the predicted $f_1(x_i;\theta)$ and actual values $y_i$. 
$\hat{\theta}^*$ is defined as the set of parameters in $\Theta$ that minimizes $\mathcal{L}$ over $p$ sequences $x_i$ with labels $y_i$ in our training dataset:
\begin{equation}
    \hat{\theta}^* = \argmin_{\theta \in \Theta} \frac{1}{p} \sum_{i=1}^p \mathcal{L}(f_1(x_i;\theta), y_i)
\end{equation}

We assume that an adversary has the ability to introduce perturbations to any point in the historical data of any client. 
Formally, we denote the $i$-th batch of historical data as $X_i = [ x_1, ..., x_m ]_i^T \in \mathbb{R}^{m \times n}$, which is composed of $m$ univariate sequences of length $n$, and the corresponding future values as $Y_i = [y_1, ..., y_m]^T_i\in \mathbb{R}^m$. 
$\delta_i \in \mathbb{R}^{m \times n}$ is the perturbation applied to $X_i$.
For each batch $i$, the aim is to find an optimal perturbation $\delta^*_i$ that maximizes the loss $\mathcal{L}$ defined in Eq.~\ref{eq:max_loss}. 
The perturbation to each sequence $x$ is limited within a $\ell_{\infty}$-norm ball of radius $\epsilon$: $\mathcal{B}_{\infty}(x, \epsilon) = \{x' \in \mathbb{R}^n : ||x' - x||_{\infty} \leq \epsilon\}$.
$\epsilon = 0.3$ means 30\% of the range of the traffic volume in the training set, where range is defined as the difference between the maximum and minimum traffic volume. 
If $X_i$ is a batch of normal data, $X_i + \delta_{i}$ is the corresponding batch of perturbed data.
Finding the optimal perturbation $\delta_i^* \in \Delta=\mathcal{B}_{\infty}(x_i, \epsilon)^m \in \mathbb{R}^{m \times n}$ for each batch $X_i$ reduces to solving the following optimization problem:
\begin{equation}
\delta^*_i = \argmax_{\delta_i \in \Delta} \mathcal{L} \left( f_1(X_i+\delta_i; \theta), Y_i \right)
\label{eq:max_loss}
\end{equation}

Here, we assume that the adversary's goal is to perturb the target forecasting model by poisoning historical data \cite{Catak2022}. 
With the ability to target any 
base station and manipulate any subsequence data, these attacks can be pervasive and difficult to be detected. 
In general, the forecaster $F_1$ can be attacked in various ways.
We choose a popular attack called Projected Gradient Descent (PGD) due to its widespread adoption in adversarial machine learning \cite{bryniarski2021evading, zhang2020attacks, liu2023robust}.

\subsection{PGD Attack}
\label{subsec:pgd_attack}

To approximate a solution for Eq.~\ref{eq:max_loss} and find the optimal disturbance that maximizes MSE of $F_1$, we use a PGD attack, as shown in Eq.~\ref{eq:pgd_iter}. 
This attack employs a projection for $T$ steps, a learning rate $\alpha$ and an initial perturbation $\delta_{i, 0} = 0 \text{ or } \delta_{i, 0} \sim \text{Uniform}(-\epsilon, +\epsilon)$. 
This approximation ultimately allows us to presume that $\delta^*_i \approx \delta_{i, T}$.
For notation purposes, we use $\tilde{X}_{i,t}$ to represent the perturbed batch $X_i$ at step $t$ with a perturbation $\delta_{i,t}$, i.e. $\tilde{X}_{i,t} = X_i + \delta_{i,t}$, with an initial perturbed batch $X_{i,0}=[ x_{1, 0}, ..., x_{m, 0}]_i^T$ as shown in Eq.~\ref{eq:x_i_0}.
\begin{equation}
\tilde{X}_{i,t+1} \leftarrow \text{Proj}_{\Delta} \left( \tilde{X}_{i,t} + \alpha \cdot \text{sign} \left( \nabla_{\tilde{X}_{i,t}} \mathcal{L}(f(\tilde{X}_{i,t}; \theta), Y_i) \right)\right)
\label{eq:pgd_iter}
\end{equation}
\begin{equation}
X_{i, 0} = X_i + \mathbbm{1}(\delta_{i, 0} \sim Uniform(-\epsilon, \epsilon)) \cdot~\delta_{i, 0}
\label{eq:x_i_0}
\end{equation}

At each step, a new perturbed batch is crafted using the direction of the loss gradient with respect to the previous perturbed batch. 
The magnitude of the perturbation is bounded by the radius $\epsilon$ of an $\ell_{\infty}$-norm ball centered on the initial input $X_{i, 0}$ as illustrated in Eq.~\ref{eq:x_i_0}.
The radius $\epsilon$ signifies the maximum permissible perturbation on the input, serving a pivotal role in our adversarial training setup. 
Each component illustrated in Figure~\ref{fig:module_training} has its unique $\epsilon$ value: $\epsilon_d$ for the denoiser D, $\epsilon_f$ for the forecaster $F_2$, $\epsilon_c$ for the classifier C, and $\epsilon_t$ during the testing phase.

In order to distinguish perturbed batches from non-perturbed ones, the classifier may require fine-tuning to endure smaller adversarial perturbations, thus a smaller $\epsilon_c$. 
On the contrary, a larger $\epsilon_f$ is essential to demonstrate robustness against substantial adversarial perturbations.
We employ a PGD attack that computes a projection within a ball of radius $\epsilon_f \geq \epsilon_c$ for the first $T-1$ steps as detailed in Eqs.~\ref{eq:delta_T_1} and \ref{eq:pgd_T-1}.
At the final step $T$, the attack yields two outputs (Eqs.~\ref{eq:delta_T} and \ref{eq:pgd_T}): one adversarial training batch for the forecaster $F_2$ within the $\epsilon_f$ radius ball and another for the classifier within the $\epsilon_c$ radius ball.
Since $\epsilon_f \geq \epsilon_t$, we project into the $\epsilon_f$ radius ball during the first $T-1$ steps. This strategy allows the perturbation to traverse a larger space before projecting into a $\epsilon_c$ radius ball, thus potentially capturing more information
For this reason, $\Delta$ depends on $t$, the projection step, so that $\Delta = \Delta_t$. We provide more information about the form of $\Delta_t$ in the following equations : 
\begin{equation}
    \forall t \in \llbracket 1, T-1 \rrbracket: \Delta_t = 
    \bigtimes^m_{i=1} \mathcal{B}_{\infty}(x_{i,0}, \epsilon_f)
    \label{eq:delta_T_1}
\end{equation}
\begin{equation}
\forall t \in \llbracket 1, T-1 \rrbracket:  \text{ Proj}_{\Delta_{t+1}}(v) = \text{Clamp}(v, x - \epsilon_f, x + \epsilon_f)
\label{eq:pgd_T-1}
\end{equation}
\begin{equation}
    \Delta_T = \Delta^f_T \times \Delta^c_T =\bigtimes^m_{i=1} \mathcal{B}_{\infty}(x_{i,0}, \epsilon_f) \times \bigtimes^m_{i=1} \mathcal{B}_{\infty}(x_{i,0}, \epsilon_c)
    \label{eq:delta_T}
\end{equation}
\begin{equation}
\text{Proj}_{\Delta_{\text{T}}}(v) = 
\begin{bmatrix}
\text{Clamp}(v, x - \epsilon_f, x + \epsilon_f)\\
\text{Clamp}(v, x - \epsilon_c, x + \epsilon_c)
\end{bmatrix}^T
\label{eq:pgd_T}
\end{equation}

Unlike the formulation in Madry et al.~\cite{madry2017towards}, $F_2$ in our work serves as a surrogate version of $F_1$. 
$F_1$ is attacked by PGD and aids in generating the poisoned samples, on which $F_2$ is trained on. 
There is no parameter sharing between $F_1$ and $F_2$, but they share the same network architecture. Both use two LSTM layers coupled with layer normalization and dropout, a fully connected layer, terminates with sigmoid activation function.
Our approach enhances the attack, since if the adversarial training and the PGD attack are performed on the same forecaster as in \cite{madry2017towards}, it naturally becomes more resistant to these attacks during training. 
Therefore, attacking it becomes more complicated, which could prevent the attack from being optimal and not reaching the boundary of the ball of radius $\epsilon$. 
Here, we attack $F_1$, which has no defense mechanism, ensuring the PGD attack retains its effectiveness throughout training.
During the testing phase, the attacker injects noise within a $\ell_{\infty}$-norm ball of radius $\epsilon_t$.

\subsection{Perturbed Sequences}
\label{subsec:sequences}

Previous works usually assume that when a client (i.e., a base station) is compromised, the entire historical sequence from that client is perturbed~\cite{liu2023robust}. 
In this work, we assume that the attacker can manipulate the value of individual time steps of each sequence from each client in order to evade detection.

To accommodate this scenario, we generate partially perturbed sequences by applying various masks to the original sequences. 
Differing from Zheng et al.~\cite{zheng_poisoning_2022} who perturb entire sequences, our method operates at a bi-level granularity.
We use two hyperparameters \%pseq to control the proportion of sequences and $k$ to determine the number of individual time-steps to perturb.
This perturbation strategy adds an extra layer of complexity and specificity in our attacks.

As an example for a sequence of length $n=3$, the mask $ q=(0, 0, 1) $ modifies the last value of a given sequence in the normal batch $X_i$ and replace it by its perturbed version, i.e., the last value of this corresponding sequence in $\tilde{X}_{i,T}$. 
For notations, we introduce $\mathbb{Q}_n$ as the set of different masks of length $n$. $|\mathbb{Q}_n| = 2^n$ for the classifier that uses both perturbed and unperturbed data, and $|\mathbb{Q}_n| = 2^n - 1$ for the denoiser and forecaster $F_2$ that use perturbed examples only, without the non-perturbation mask $(0, 0, 0)$.
Each mask $q$ is a binary sequence that can be drawn uniformly from $\mathbb{Q}_n$ to perturb sequences in the batch. 
Applying this mask implies replacing a value in a sequence by its perturbed version if and only if the mask at the corresponding position takes a value 1. 
The final masked, perturbed batch takes the form $X_i + q \odot \delta_{i, T}$, as illustrated in Eq.~\ref{eq:mask}.
We utilize the Hadamard product, denoted by $\odot$, such that $q \odot A$ corresponds to the element-wise multiplication of each row of $A$ by each element of $q$, resulting in a matrix of the same shape as $A$.
\begin{equation}
(\mathbbm{1}_n-q) \odot X_i + q \odot \tilde{X}_{i,T} = X_i + q \odot \delta_{i, T}
\label{eq:mask}
\end{equation}

\subsection{Attacker Capabilities and Knowledge}
\label{subsec:capa}

While previous works assume that the adversaries have limited knowledge and abilities, we assume in this work that an attacker is very familiar with wireless traffic forecasting systems and can manipulate the data at will.
In particular, this adversary can target any, and possibly multiple, base stations and individual data points using a $T=10$-steps PGD attack with the following knowledge: 
\begin{itemize}
    \item The adversary needs to know the targeted forecaster, including its architecture and the associated weights, biases, and loss function. This is because a PGD attack involves tweaking the input data to make the forecaster perform poorly, which requires understanding how it works. 
    \item The adversary needs to know about the data used to train the targeted forecaster. This is because a PGD attack involves creating misleading examples based on the training data to trick the forecaster.
\end{itemize} 

The assumed capabilities allow us to have a robustness analysis under the most challenging conditions, i.e., adversary has complete information as that of defender, which is known as a white box attack.
One may argue that such a sophisticated attack may not be feasible in reality. 
 
We, however, take this “worst-case scenario'' as a stress test for any vulnerable deep forecasting model. 
In other words, if a model can withstand an attack by such a well-equipped attacker, its robustness and resilience could be sufficiently validated.

It is worth noting that the novelty of our proposed attack lies in the adversary's ability to manipulate numerous base stations, potentially simultaneously, at time steps of their choice. 
This significantly amplifies the potential strength of the attack compared to those previous attacks studied in the field, such as the attacks against Bayesian forecasting dynamic models \cite{naveiro2021adversarial}, and the targeted attacks to time series forecasting models \cite{govindarajulu2023targeted}. 
Similarly, the attacks on multivariate probabilistic forecasting models reported in \cite{liu2023robust} did not consider manipulating numerous base stations at chosen time steps. 
By considering a more powerful adversary, we extend the previous studies about the robustness of deep forecasting models under more sophisticated attack scenarios, which will benefit the design of more resilient wireless network prediction systems.

\section{Defense Mechanisms}
\label{sec:defense}

\begin{figure}[tb]
    \centering
    \includegraphics[width=0.4\textwidth]{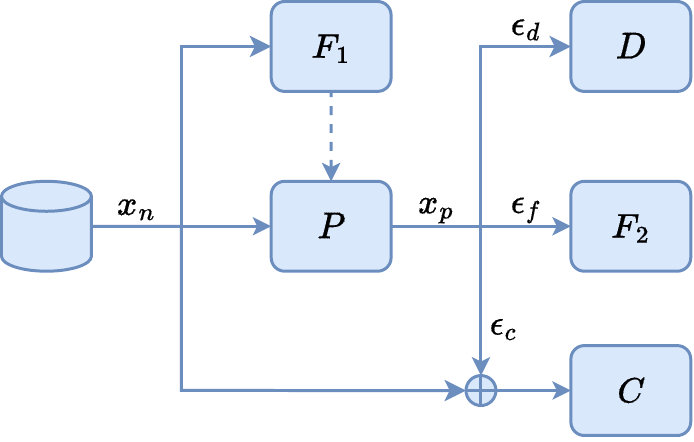}
    \caption{Four components trained separately: $F_1$ is the forecaster trained only on normal samples $x_n$ and used for PGD attack $P$, which produces adversarial samples $x_p$. 
    The classifier $C$ is trained on a mix of normal and poisoned samples, whereas the denoiser $D$ and the forecaster $F_2$ are trained on adversarial samples only.}
    \label{fig:module_training}
\end{figure}

\begin{figure*}[tb]
    \centering
    \includegraphics[width=0.85\textwidth]{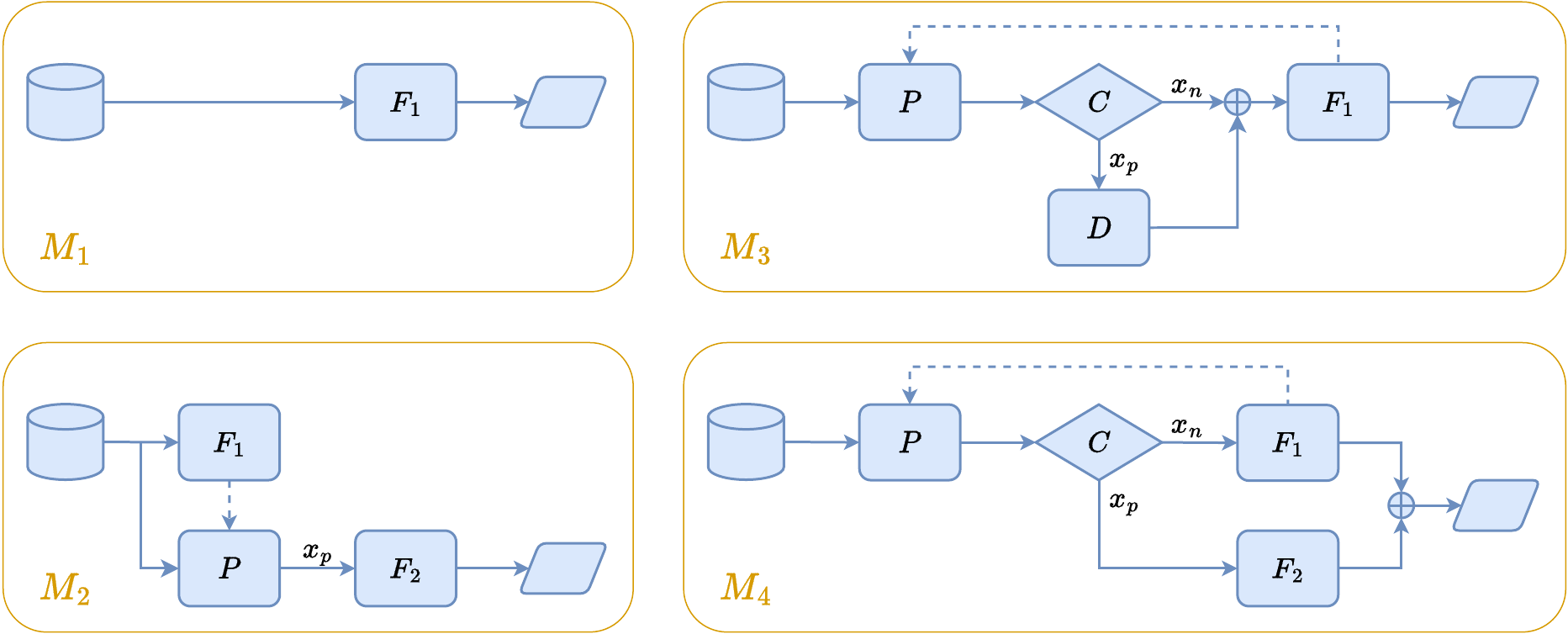}
    \caption{The four models and their composing components presented in Fig.~\ref{fig:module_training}.
    $M_1$ and $M_2$ employ forecasters $F_1$ and $F_2$. $M_3$ utilizes classifier $C$ (adversarial sample detector) and denoiser $D$ (if perturbation detected), followed by forecaster $F_1$. $M_4$ uses classifier $C$ with $F_2$ (if perturbation detected), or with $F_1$ otherwise. $x_n$ and $x_p$ represent normal and perturbed sequences, respectively.}
    \label{fig:evaluation_frameworks}
\end{figure*}

To cope with the given attacks, we develop our defense mechanisms in a systematic approach by leveraging and integrating the components in Fig.~\ref{fig:module_training}. These defense mechanisms, called models $M_i$, are shown in Fig.~\ref{fig:evaluation_frameworks}. In particular, we include two baseline models $M_1$ and $M_2$ for the comparison purpose. 
As shown in the figure, $M_1$ and $M_2$ make use of the two forecasters $F_1$ (trained with normal samples) and $F_2$ (trained with $100\%$ poisoned samples). $M_3$ is composed of classifier $C$ that detects adversarial samples and denoiser $D$ that removes perturbation from the detected adversarial samples, along with forecaster $F_1$. 
In doing so, we transform the problem of model's performance and robustness to adversarial attacks into adversarial example classification and feature denoising, breaking the traditional trade-off between model robustness and accuracy. 
$M_4$ uses classifier $C$ with forecaster $F_1$ if the $C$ considers the sequence as normal, or with forecaster $F_2$ otherwise.
In our approach, the components $F_1$, $F_2$, $C$, and $D$ are trained in an independent manner from one another. Thanks to this structure, the performance of individual components does not have an impact on each other. It is also computationally efficient since components can be trained in parallel. 

\subsection{Adversarial Training}
\label{subsec:adv_train}

Adversarial training, a method for bolstering machine learning models against adversarial attacks, involves training on a blend of clean and perturbed examples to ensure model performance even when the instances are manipulated by an adversary\cite{madry2017towards, zhu2018optimal, wen2019embedding}. 
This approach can be computationally costly and requires a trade-off between accuracy on clean and adversarial examples. 
To mitigate these issues, recent work embeds adversarial perturbations into the neural network's parameter space \cite{wen2019embedding}, while alternative methods such as label smoothing and logit squeezing mimic adversarial training mechanisms \cite{shafahi2019label}.

The mathematical distinction between Empirical Risk Minimization (ERM) and Adversarial Risk Minimization (ARM) lies in their objective functions. 
In particular, ERM minimizes empirical loss over the data distribution (Eq.~\ref{eq:ERM}), optimizing for average-case performance, while ARM minimizes the maximum loss over all possible perturbations $\delta$ within a set $\Delta$ (Eq.~\ref{eq:ARM}), optimizing for worst-case performance. 
This makes ARM more computationally demanding than ERM, as it requires solving an additional optimization problem for each subsequence in order to find the loss-maximizing perturbation $\delta$.
\begin{equation}
\hat{\mathcal{R}}(\theta) = \frac{1}{m} \sum_{i=1}^{m} \mathcal{L}(f(X_i; \theta), Y_i)
\label{eq:ERM}
\end{equation}
\begin{equation}
\hat{\mathcal{R}}_{\text{adv}}(\theta) = \frac{1}{m} \sum_{i=1}^{m} \max_{\delta_i \in \Delta} \mathcal{L}(f(X_i + q \odot \delta_{i, T}; \theta), Y_i)
\label{eq:ARM}
\end{equation}

Fig.~\ref{fig:module_training} which shows that the two forecasters $F_1$ and $F_2$ share identical architectures but differ in their training regimens. 
Specifically, $F_1$ undergoes standard training and serves as the target for a PGD attack to generate adversarial examples, which are then utilized to train the counterpart component, $F_2$.
The unique aspect of $F_2$ is its training on adversarial data, aiming to enhance its robustness against such attacks. 
It plays a crucial role in our defense model $M_4$, providing prediction of poisoned sequence predicted by the classifier $C$. 
Similarly, the $M_3$ model is using adversarial examples through $C$ and $D$ trainings.

It is worth noting here that despite its architectural similarity to $F_1$, $F_2$ requires an extended training period to converge due to the adversarial nature of its training (100\% perturbed samples are used for training). 
This extended training period is necessary for enhanced robustness against significant perturbations.

\subsection{Classification of Adversarial Examples}
\label{subsec:classif}

The purpose of classification of adversarial examples in time series data is to identify whether a given sequence has been perturbed or not. 
This can be essentially treated as a traditional anomaly detection problem, which aims at pinpointing anomalous sequences in their entirety \cite{yu2018asp, deepant, alemany2020dilemma}. 
While anomaly detection flags deviations from normal behavior, adversarial examples entail stealthily crafted perturbations to deceive the target models.

In our methodology, we adopt a balanced training regimen for $C$, e.g., 50\% normal sequences and 50\% perturbed sequences with norm value $\epsilon_c$ in each batch (Fig.~\ref{fig:module_training}). 
The classifier has binary output: $0$ for unperturbed sequences and $1$ for sequences with at least one perturbed time step. 

$C$ employs a series of Inception modules, each composed of 1D convolutions with kernel sizes of 1 and 3 and padding equal to 0 and 1 respectively. 
Following each Inception module, batch normalization is performed, followed by ReLU activation and dropout. 
Afterward, Global Average Pooling is conducted before a final fully connected layer. 
The loss is computed using cross-entropy.

By classifying each sub-sequence, our aim is to detect even the most subtle perturbations within the time series data. Even if only a single time step is perturbed, we flag the entire sequence as adversarial. This approach poses a unique challenge for classification, as it demands the detection of perturbed sequences even when the perturbation is minimal. However, this heightened sensitivity to minor perturbations enhances our system's detection capabilities, thereby fortifying the defense mechanism against sophisticated adversarial attacks.

\subsection{Denoising Poisoned Data}
\label{subsec:denoising}

The denoiser is designed to remove poisoned samples and make the sequences clean. 
It undergoes training with entirely noisy sequences with norm value $\epsilon_d = \epsilon_f$, with labels mirroring these sequences without noise (Fig.~\ref{fig:module_training}). 
Within the $M_3$ model, the denoiser serves to expunge noise from sequences flagged as noisy by the classifier, as depicted in Fig.~\ref{fig:evaluation_frameworks}.  
Following the denoiser's application, the normal forecaster $F_1$ is employed, considering the denoiser's output sequences as normal. 

This component is pivotal in maintaining the integrity of the base data within the training dataset.
In the event of new data being disturbed, the denoiser can be employed to remove the noise.
This ensures that the original training dataset remains free of noisy data.

The Denoiser $D$ employs a basic auto-encoder architecture.
The encoder part uses of a linear layer, batch normalization, ReLU activation, and dropout, followed by a decoder that mirrors the same structure. 
It employs MSE as its loss function.

\section{Experimental Evaluation}
\label{sec:exp}

\subsection{Dataset Description}
\label{subsec:dataset_description}

In our time series forecasting study, we use the Telecom Italia dataset \cite{barlacchi_telecom_italia_2015} as a comprehensive source of information. 
Specifically, we focus on call volumes in Milan's urban environment. 
This dataset provides a rich and detailed source of information, allowing us to delve deep into the patterns and trends of telecommunications usage. 
To align with a prior study by Zheng et al.~\cite{zheng_poisoning_2022}, in our research, we choose 100 base stations using a consistent random seed.
These base stations provide a broad and representative sample of the telecommunications activity in the city. 
Our study is analyzing hourly data over an 8-week period, 7 of which are used for training and 1 for testing. 
In order to predict time $t$, we divide the univariate time series for each base station into sub-sequences of length 3 ($t-1$, $t-2$, and $t-24$ hours). 
This approach allows us to capture the temporal dependencies in the data, which is crucial for accurate forecasting. 
However, defending against adversarial attacks is particularly challenging due to the dataset's high standard deviation, with an average standard deviation of $0.26$ after 0-1 normalization. 
This high-level variation in the data makes it difficult to distinguish normal fluctuations from adversarial perturbations, presenting a significant challenge in our defense against adversarial attacks.

\subsection{Component Updates}
\label{subsec:model_updates}

In our study, we adopt a unique approach for training that strikes a balance between computational efficiency and stability of updates. 
Specifically, we iterate over the batches within each epoch, compute the gradients for each batch, but only update the component's parameters after each epoch. 
This method is akin to batch gradient descent with a batch size equal to the size of our dataset, and it offers at least two advantages, 
\begin{itemize}
\item This method ensures that each update is based on a comprehensive view of the data, reducing the influence of outliers and noise on the learning process, thereby enhancing the stability of our updates. 
By aggregating the gradients over the entire dataset before performing an update, we mitigate the risk of erratic component behavior that can arise from the high variance in our data. 
\item It is computationally efficient. 
While benefiting from the granularity of batch-wise gradient computation, it avoids the computational cost of frequent backpropagation steps associated with updating the component parameters after each batch. 
\end{itemize}
 
In particular, $F_1$, $F_2$, $C$ and $D$ are trained independently, each epoch of which involves processing batches with a length $512$.

\subsection{Evaluation Metrics}
\label{subsec:eval_metrics}

Our objective is to have the MSE of our hybrid models $M_3$ and $M_4$ as close as possible to the MSE of our standard model $M_1$ on unperturbed data.
We can consider one hypothetical scenario with two models, $M_1$ and $M_3$. 
Let's imagine that initially, $M_1$ performs better than $M_3$ on standard, unperturbed data, with an MSE of 1.5 versus 2. 
However, when we introduce data perturbations, the narrative can shift significantly.
$M_3$ might show remarkable resilience to these data perturbations, maintaining, for instance, 90.9\% (e.g., an increase from 2 to 2.2 MSE) of its original performance level, indicating its superior adaptability to challenging conditions. 
Conversely, $M_1$'s performance could deteriorate dramatically, only retaining around 33\% (an increase from 1.5 to 4.5 MSE) of its original effectiveness.
This example emphasizes that even if a model seems slightly less effective on clean data, its ability to retain a high percentage of its original performance in the face of real-world perturbations could make it the preferred choice for handling unpredictable data scenarios.

In addition to evaluating the robustness of our models based on their MSE, we are also interested in the accuracy of $C$ in detecting adversarial examples, as it provides insight into how challenging it is for the system to identify these intentionally misleading data points. 
We examine this metric under various parameters and consider how it changes with different levels of data perturbation.
%, or how it performs with different $\epsilon$-norms values. 
By analyzing the classifier's accuracy in this way, we can gain a more comprehensive understanding of our system's resilience to adversarial attacks and identify potential areas for improvement.

\subsection{Experimental Setup}
\label{subsec:experimental_setup}

Our setup involves historical data with a length $n=3$.
Each time step in the subsequence can either be perturbed or not, leading to one normal version and $7$ possible perturbed versions. 
This approach is more comprehensive than other methods that perturb the entire sequence \cite{madry2017towards, zhu2018optimal, wen2019embedding}. 
By using this perturbation approach, we can gain a better understanding of the potential vulnerabilities of our forecasting $F_1$ and train a more robust adversarial forecaster $F_2$. 
This is because we consider a wider range of attacks, compared to fully perturbed sequences.

\subsubsection{Components}
\label{subsubsec:compo}

Four components are trained: a Forecaster $F_1$ and his surrogate $F_2$, a Denoiser $D$ , and a Classifier $C$.
These components are then assembled at test time to form the four models.
All of them are implemented using PyTorch and trained using the Adam optimizer with a batch size of $512$.

All the corresponding hyperparameters can be found in Table ~\ref{tab:hyper}.
We trained $F_1$ for 10 epochs as in \cite{zheng_poisoning_2022} and the adversarial version $F_2$ over 15 epochs, due to the difficulty of reaching convergence in adversarial training.
The two remaining components $C$ and $D$ are trained over 40 epochs, a good balance to learn representations without overfitting the training noise.

\begin{table}[ht]
    \centering
    \caption{Hyperparameters used for components training }
    \label{tab:hyper}
    \begin{tabular}{lcccc}
        \toprule
        \multirow{2}{*}{Parameter} & \multicolumn{4}{c}{Models} \\
        \cmidrule{2-5} 
         & \textbf{$F_1$} & \textbf{$F_2$} & \textbf{$C$} & \textbf{$D$} \\
        \midrule
        \#training epochs & 10 & 15 & 40 & 40 \\
        % \hline
        Training perturbation ($\ell_{\infty}$) & 0 & $\epsilon_f$ & $\epsilon_c$ & $\epsilon_d$ \\
        % \hline
        Learning rate & 0.008 & 0.008 & 0.01 & 0.005 \\
        % \hline
        Weight decay & 0.2 & 0.2 & 0.02 & 0.1 \\
        % \hline
        Gamma & 0.5 & 0.5 & 0.5 & 0.5 \\
        % \hline
        Scheduler step size & 5 & 5 & 10 & 5 \\
        \bottomrule
    \end{tabular}
\end{table}

% \begin{table}[tb]
%     \centering
%     \caption{Performance of the four models on the test data without perturbation $(\epsilon_t = 0)$ under two training conditions ($\epsilon_c, \epsilon_f$).}
%     \begin{tabular}{ccc}
%         \toprule
%         \multirow{2}{*}{Model} & \multicolumn{2}{c}{MSE} \\
%         \cline{2-3} 
%          & $(\epsilon_c, \epsilon_f) = (0.3, 0.3)$ & $(\epsilon_c, \epsilon_f) = (0.2, 0.3)$  \\
%         \midrule
%         M1 & 0.0173 & 0.0173 \\
%         M2 & 0.0509 & 0.0509 \\
%         M3 & 0.0190 & 0.0188 \\
%         M4 & 0.0257 & 0.0234 \\
%         \bottomrule
%     \end{tabular}
%     \label{tab:mse_normal}
% \end{table}

\subsubsection{Models}
\label{subsubsec:models}

The four models are assembled and used only for inference. 
They are composed of one or more previously trained components, as depicted in Fig.~\ref{fig:evaluation_frameworks}.
We evaluate the performance of these models.

\subsubsection{Objectives}
\label{subsubsec:obj}

Models engineered to be robust to noise prove their worth when applied to noisy sequences, thereby outperforming non-robust models. 
However, this robustness comes at a cost: when these models are applied to non-noisy sequences, their performance often declines compared to non-robust models. 
This presents a significant problem : if the sequences are typically non-noisy, the usage of a robust model could lead to substantial loss in accuracy.

Our objective is to maintain the performance level of a non-robust model on non-disturbed sequences, while reaping the benefits of robust models when applied to disturbed sequences.

\subsubsection{Evaluation}

In evaluating the models, their resistance to the PGD attack is assessed by comparing their performance on clean data and perturbed data, and serves as a measure of the models' robustness against adversarial perturbations. 

To explore the models' resistance to different levels of perturbations, we vary two parameters: the number of perturbed steps, denoted as $k$, and the percentage of perturbed sequences in the test set, denoted as \%pseq. 
For $k$, we examine the range from 0 (no perturbation) to 3 (all steps perturbed). 
Regarding \%pseq, we investigate four scenarios: 0\% (no perturbation), 20\%, 50\%, and 100\% perturbed sequences in the test set. 
These settings allow us to assess the models' performance across a spectrum of perturbation levels, ranging from zero disturbance with normal data up to the entirety of the testing set being disturbed.

During the evaluation, we examine the impact of varying the $\epsilon$-norm values in the triplet $(\epsilon_c, \epsilon_f, \epsilon_t)$ by setting $\epsilon_d = \epsilon_f$.
Specifically, we observe how the models react when the testing phase's $\epsilon$-norm value, $\epsilon_t$, differs from the values of $\epsilon_c$ and $\epsilon_f$.

Through this evaluation, we aim to demonstrate that there is a distinct advantage in decoupling $\epsilon_c$ and $\epsilon_f$ during training.
Furthermore, we aim to show that a smaller $\epsilon_c$ allows the classifier to segregate data with $\epsilon_t \geq \epsilon_c$ during inference.
Similarly, we seek to illustrate that a larger $\epsilon_f$ exhibits resilience at inference, even if $\epsilon_t \leq \epsilon_f$.

\subsection{Results}
\label{subsec:robustness_analysis}

We compared a baseline model $M_1$, a robust baseline model $M_2$, and two hybrid variants $M_3$ and $M_4$ (our defense mechanisms) as shown in Fig.~\ref{fig:evaluation_frameworks}.
The first variant $M_3$ is a model that adds two extra components: the classifier $C$ and the denoiser $D$. 
If the classifier predicts a sequence is noisy, the denoiser is used to remove the noise, and the non-robust forecaster $F_1$ is subsequently applied. 
If no disturbance is detected, $F_1$ is applied directly. 
The second variant $M_4$ involves the classifier that predicts whether a sequence is disturbed. 
If a disturbance is detected, the robust forecaster $F_2$ is applied; otherwise, $F_1$ is used.

We select 100 base stations to predict call volume in Milan, incorporating data points from 1 hour, 2 hours, and 24 hours prior to the prediction, testing hourly for a week following a 7-week training period. 
The results in the tables represent the average MSE across all base stations for the test week.
All results presented are from the testing phase.

Both $F_2$ and $D$ were trained with a disturbance norm $\epsilon_{f} = \epsilon_{d}=0.3$ throughout the training process. 
However, the training norm $\epsilon_{c}$ for $C$ is varied from $0.2$ to $0.3$, and we also observe test results with different norms $\epsilon_{t}$ from $0.1$ to $0.4$.

\subsubsection{Clean data with $\epsilon_c = \epsilon_f$}
\label{subsubsec:clean_dat_equal}

\begin{table}[tb]
    \centering
    \caption{Classifier accuracy on the test data without perturbation $(\epsilon_t = 0)$ under two training conditions ($\epsilon_c, \epsilon_f$).}
    \begin{tabular}{cc}
        \toprule
        % \multicolumn{2}{c}{Accuracy} \\
        % \cmidrule{1-2} 
         $(\epsilon_c, \epsilon_f) = (0.3, 0.3)$ & $(\epsilon_c, \epsilon_f) = (0.2, 0.3)$  \\
        \midrule
        60.93\% & 61.23\% \\
        \bottomrule
    \end{tabular}
    \label{tab:acc_normal}
\end{table}

\begin{table}[tb]
    \centering
    \caption{Performance of the four models on the test data without perturbation $(\epsilon_t = 0)$ under two training conditions ($\epsilon_c, \epsilon_f$).}
    \begin{tabular}{ccc}
        \toprule
        \multirow{2}{*}{Model} & \multicolumn{2}{c}{MSE} \\
        \cmidrule{2-3} 
         & $(\epsilon_c, \epsilon_f) = (0.3, 0.3)$ & $(\epsilon_c, \epsilon_f) = (0.2, 0.3)$  \\
        \midrule
        M1 & 0.0173 & 0.0173 \\
        M2 & 0.0509 & 0.0509 \\
        M3 & 0.0190 & 0.0188 \\
        M4 & 0.0257 & 0.0234 \\
        \bottomrule
    \end{tabular}
    \label{tab:mse_normal}
\end{table}

% \begin{table*}[tb]
% \centering
% \begin{tabular}{|c|c|}
% \hline
% \textbf{Model} & \textbf{MSE without any perturbation} \\
% \hline
% M1 & 0.0173\\
% \hline
% M2 & 0.0509\\
% \hline
% M3 & 0.0190 \\
% \hline
% M4 & 0.0258 \\
% \hline
% \end{tabular}
% \caption{Performance of the models on normal samples under training conditions with poison level of 0.3 and no perturbation in the test}
% \label{tab:model_normal_perf}
% \end{table*}

% \begin{table*}[tb]
% \centering
% \begin{tabular}{|c|c|}
% \hline
% \textbf{Model} & \textbf{MSE without any perturbation} \\
% \hline
% M1 & 0.0173\\
% \hline
% M2 & 0.0509\\
% \hline
% M3 & 0.0188 \\
% \hline
% M4 & 0.0234 \\
% \hline
% \end{tabular}
% \caption{Summary of model MSE losses under training conditions with $\epsilon_{c}=0.3$, $\epsilon_{f}=0.3$ and testing conditions with a poison level of $\epsilon_{t}=0.3$}
% \label{tab:mse_no_perturb_0.3_0.2}
% \end{table*}

The results presented in Table~\ref{tab:mse_normal} demonstrate a significant improvement in the effectiveness of $M_3$ and $M_4$, particularly when there is no poisoning involved, compared to $M_2$~\cite{madry2017towards}. 
%When compared to a model that is trained solely on adversarial examples, such as $M_2$, our models $M_3$ and $M_4$ show a notable improvement \cite{madry2017towards}.
Specifically, while the robust $M_2$ model experiences an MSE multiplied by a factor of 2.94 (from 0.0173 to 0.0509) on non-poisoned sequences, $M_3$ and $M_4$ manage to maintain a performance almost equivalent to that of model $M_1$ with an MSE increasing only by a factor of 1.1 and 1.49 respectively.

Table~\ref{tab:acc_normal} shows the classifier's accuracy on normal samples. 
60.93\% of the normal sequences are correctly classified, thus 39.07\% of are incorrectly predicted as perturbed. 
This high false positive rate is likely due to the high natural variance in the original dataset. 
%However, we speculate in datasets with lower natural variance, the classifier's results should improve.
Interestingly, with a low accuracy of 60.93\%, $M_3$ manages to match the base MSE of model $M_1$. 
This suggests that there is room for further improvement. 
In fact, out of $M_3$'s MSE of 0.0190, 91.1\% is attributed to the base error of model $M_1$, and 8.9\% is due to the additional error of both the classifier and the denoiser.
We can assume that if this additional error tends towards zero, then the MSE of $M_3$ on perturbed data would be equal to that of $M_1$ on normal data. 
This would effectively break the trade-off between robustness and accuracy.
%Having established that model $M_3$, in particular, can retain nearly all of model $M_1$'s performance on non-poisoned data, 

We now turn our attention to how these models perform when faced with poisoned data.

\subsubsection{Perturbed data with $\epsilon_c = \epsilon_f$}
\label{subsubsec:p_data_equal}

Table~\ref{tab:mse_loss_perturbed} shows the results as a function of $k$ and \%pseq. 
We note that our $M_3$ model performs significantly better than the $M_2$ model with an MSE up to 2.33x lower.
%
%I 
The exception is when the whole testing set is perturbed.
However, to achieve this result, we would need to assume that an attacker could perturb all 100 base stations at each time step, a scenario that is practically implausible.

We also examine the classifier's ability to detect perturbed sequences, as shown in Table~\ref{tab:classifier_accuracy}. 
We observe high accuracy when $k$ is either 1 or 2 and \%pseq is either 20\% or 50\% but notice a decrease in accuracy in two specific scenarios. 
Firstly, when $k$ increases, it becomes more challenging for the classifier to detect the poison because this trained neural network may rely on certain metrics, such as the difference in values between two points or the standard deviation whereas if $k=3$, most of the sequences simply shifts in level, making the detection more complicated.
The second scenario is when $\epsilon_t \leq \epsilon_c$, as it is more difficult for the model to separate noisy and normal sequences with a smaller $\epsilon_t$.

Given these observations, we plan to test a new scenario by decoupling $\epsilon_c$ and $\epsilon_f$, specifically, setting $\epsilon_c$ to be less than $\epsilon_f$. 

\subsubsection{Clean data with $\epsilon_c < \epsilon_f$}
\label{subsubsec:clean_data_lessthan}

We first observe that decoupling $\epsilon_c$ and $\epsilon_f$ slightly improves performance. 
More than the raw results of accuracy or MSE, we are interested in their correlation when $\epsilon_c$ goes from $0.3$ to $0.2$.
We notice that a 0.5\% increase in classifier accuracy corresponds to a 1.1\% decrease in $M_3$'s MSE and an 8.9\% decrease in $M_4$'s MSE. 
Given that the base error of the model is 0.0173, the relationship between accuracy increase and MSE decrease is likely non-linear, with the MSE's decrease diminishing as accuracy becomes very high.
This correlation is noteworthy, and the precise relationship between accuracy's increase and MSE's decrease in this context could be a subject for future research.
In comparison to the clean data when $\epsilon_c = \epsilon_f$, $M_3$ recovers 92.02\% of the baseline error from $M_1$, with an MSE 2.71x lower than $M_2$.
$M_4$ is now recovering $73.93\%$ of $M_1$'s error, representing an increase of 6.61\% compared to the previous scenario.

\subsubsection{Perturbed data with $\epsilon_c < \epsilon_f$ and $\epsilon_t \leq \epsilon_f$}
\label{subsubsec:p_data_lessthan}

We examine results for $\epsilon_c=0.2$, $\epsilon_f=0.3$, and $\epsilon_t$ values varying from 0.10, to 0.30. 
%M_1$ outperforms the other models when $k=1$.
%$M_2$ outperforms only when $k=3$, \%pseq=100\% and $\epsilon_t \geq \epsilon_c$, as previously mentioned.
In most of the scenarios, $M_3$ performs best.
%We also observe trends as $\epsilon_t$ increases, considering the average case where $k=2$ and \%pseq=50\%. As $\epsilon_t$ increases from 0.1 to 0.3, $M_1$'s MSE rises from 0.0230 to 0.0434. 
%Meanwhile, $M_2$'s MSE increases from 0.0517 to 0.0532, $M_3$'s from 0.0229 to 0.0377, and $M_4$'s from 0.0334 to 0.0391.
Several observations can be made from the results. 
Firstly, without adversarial training, $M_1$'s MSE significantly increases when facing perturbations. 
Secondly, $M_2$'s MSE experiences the smallest increase, but this MSE is much higher compared to $M_3$ and $M_4$.
Finally, we note that the benefit of robust models becomes limited when $\epsilon_t$ is very small. 
In this scenario, $M_1$ gains ground on the others, suggesting adversarial training is beneficial when $\epsilon_t \geq 0.1$. 
$M_2$ is suitable for extreme cases where the entire test can be perturbed with very large $\epsilon_t$, exceeding $0.3$, which is challenging in practice. 
Therefore, we prefer the use of $M_3$.

Finally, regarding the classifier, we aim to address our hypothesis: if a classifier is trained to detect adversarial examples with $\epsilon_c$, does this imply it can detect them with $\epsilon_t$ greater than $\epsilon_c$? 
The answer is yes, except for the unique case of a shift of the whole sequence for $k=3$.
%, as seen by comparing accuracies for (0.2, 0.3, 0.3) and (0.2, 0.3, 0.2).
%As previously discussed, this case is unique, as the entire sequence is altered. 
%For instance, if the PGD attack adds $+0.3$ perturbation to each step, it merely changes the level.
Lastly, we also observe that as $\epsilon_t$ becomes small, it becomes increasingly challenging for the classifier to separate the data.

\subsubsection{Perturbed data with $\epsilon_c < \epsilon_f$ and $\epsilon_t > \epsilon_f$}
\label{subsubsec:p_data_equal_greater}

Finally, when $\epsilon_t$ is larger than $\epsilon_c$ and $\epsilon_f$, model $M_4$ appears to be a good compromise for $k=2$ and \%pseq $\leq 50\%$. 
For $k=3$ and \%pseq $\geq 50\%$, we would prefer model $M_2$ with adversarial training. 
However, we can assume that in such a case, experts could easily detect perturbations with $\epsilon_t=0.4$ on all the historical data steps for each of the 100 base stations. 

For the classification task, the performance is quite similar to (0.2, 0.3, 0.3).

\subsection{Discussion and Comparison}

One of our key findings is the potential of model $M_3$, composed of $3$ components: a classifier to detect the poisoned samples, a denoiser to remove the noise from them, and a forecaster trained on normal samples only. 
Specifically, $M_3$ performed $2.71\times$ better than $M_2$, retaining 92.02\% of $M_1$'s MSE on normal data, while maintaining performance on perturbed data that was up to $1.71\times$ lower than $M_1$, especially in the cases of high perturbation, and $2.51\times$ lower than $M_2$, particularly in the cases of low perturbation.
We therefore postulate that if the classifier approaches an accuracy near 100\% and the denoiser's MSE approaches $0$, the performance of $M_3$ on perturbed data would align with the performance of the standard model $M_1$ on clean data. 
This effectively breaks the traditional trade-off between robustness and accuracy.

Furthermore, our results confirmed that a classifier trained to detect adversarial examples with a certain $\epsilon_c$-norm value can detect them if $\epsilon_t \geq \epsilon_c$ at test time, except when all the steps in historical data are perturbed. 
%This finding further supports the effectiveness of our proposed %approach.
The results showed significant difference from the ones reported in~\cite{zheng_poisoning_2022}, which employed an LSTM-based methodology for predicting call volume, while the performance was reduced down to $72.39\%$ of the original model's performance if defense mechanism was incorporated. 
This resulted in an MSE of $0.0902$, marking an increase of $1.38\times$ from their original MSE value of $0.0653$. 
On the other hand, our  $M_3$ showcased enhanced resilience, maintaining 92.02\% of the baseline performance after defense, resulting in an MSE rising only by a factor of $1.09$.
The comparative evaluation underscores the efficiency of our models (especially of \(M_3\)) on mitigating the ramifications of adversarial attacks, and safeguarding the fidelity of time series forecasting.

\vfill\eject
\section{Conclusion}
\label{sec:conclusion}

In this paper, we addressed the long-standing challenge about simultaneously achieving performance and robustness of deep forecasting models in the face of poisoning attacks.  
We systematically investigated the strengths and weaknesses of each model under different scenarios by varying the perturbation levels and $\epsilon$-norm values, and studied the complementary roles of adversarial training, adversarial examples classification, and denoising in enhancing model robustness. 

We took wireless traffic prediction as a specific scenario and developed a new type of attack where an attacker can perturb any subsequence step of any base station with a 10-steps PGD attack on clean forecaster $F_1$.
We then proposed two defense mechanisms by assembling adversarially (versus cleanly) trained forecaster, classifier, and denoiser, with the objective to maintain performance on normal data while preserving robustness to poisoned data. 
By theoretically and experimentally demonstrating the possibility of breaking the trade-off between model robustness and accuracy, we believe that our findings laid down a foundation for further development of robust and reliable deep forecasting models, which therefore deserve more attention and effort from the community.

\begin{landscape}
\begin{table}[t]
    \centering
    \caption{Classifier accuracy (in \%) based on the percentage of perturbed sequences (\%pseq) and the number of perturbed steps $k$ for different values of poison levels for classifier ($\epsilon_c$) and forecaster ($\epsilon_f$) in training and testing ($\epsilon_t$).}
    \label{tab:classifier_accuracy}
    \setlength{\tabcolsep}{4.4pt}
    \begin{tabular}{cccccccccccccccccccccccc}
        \toprule
        \multirow{3}{*}{\%pseq} 
         & \multicolumn{21}{c}{Classifier accuracy at different $(\epsilon_c, \epsilon_f, \epsilon_t)$} \\
        % \cline{2-19} 
        \cmidrule(rl){2-22}
         & \multicolumn{3}{c}{$(0.3, 0.3, 0.3)$} 
         & \multicolumn{3}{c}{$(0.3, 0.3, 0.2)$}
         & \multicolumn{3}{c}{$(0.2, 0.3, 0.3)$} 
         & \multicolumn{3}{c}{$(0.2, 0.3, 0.2)$} 
         & \multicolumn{3}{c}{$(0.2, 0.3, 0.15)$} 
         & \multicolumn{3}{c}{$(0.2, 0.3, 0.1)$}
         & \multicolumn{3}{c}{$(0.2, 0.3, 0.4)$}\\
        % \cline{2-19} 
        \cmidrule(rl){2-4} \cmidrule(rl){5-7} \cmidrule(rl){8-10} \cmidrule(rl){11-13} \cmidrule(rl){14-16} \cmidrule(rl){17-19} \cmidrule(rl){20-22}
         & $k=1$ & $k=2$ & $k=3$ & $k=1$ & $k=2$ & $k=3$ & $k=1$ & $k=2$ & $k=3$ 
         & $k=1$ & $k=2$ & $k=3$ & $k=1$ & $k=2$ & $k=3$ & $k=1$ & $k=2$ & $k=3$ 
         & $k=1$ & $k=2$ & $k=3$\\ 
        \midrule
        20  & 80.78 & 79.28 & 58.75 & 76.12 & 75.14 & 62.32 & 77.42 & 75.61 & 55.98 
            & 74.81 & 73.83 & 64.27 & 72.42 & 72.10 & 65.07 & 69.03 & 69.17 & 60.34 & 78.32 & 75.88 & 51.68 \\ 
        50  & 86.64 & 83.85 & 60.48 & 82.70 & 80.31 & 51.42 & 82.62 & 80.68 & 61.26 
            & 80.31 & 79.33 & 66.88 & 77.35 & 77.11 & 62.49 & 71.35 & 72.48 & 54.49 & 83.02 & 80.16 & 44.74\\ 
        100 & 70.52 & 67.64 & 31.70 & 66.47 & 62.58 & 33.97 & 67.64 & 68.99 & 56.51 
            & 64.15 & 65.01 & 41.29 & 60.82 & 58.44 & 39.34 & 54.39 & 49.78 & 39.60 & 69.15 & 70.39 & 54.62 \\
        \bottomrule
    \end{tabular}
% \end{table}
%

% \hspace*{2cm}
% \bigskip
\vspace*{3cm}

%
% \begin{table}[tb]
    \centering
    \caption{Performance of the four models at different values of the percentages of perturbed sequences (\%pseq), the number of perturbed steps $k$, and the poison levels for classifier ($\epsilon_c$) and forecaster ($\epsilon_f$) in training and testing ($\epsilon_t$).}
    \label{tab:mse_loss_perturbed}
    \setlength{\tabcolsep}{1.2pt}
    \begin{tabular}{ccccccccccccccccccccccc}
        \toprule
        \multirow{3}{*}{\%pseq} & \multirow{3}{*}{Model}
         & \multicolumn{21}{c}{MSE at different $(\epsilon_c, \epsilon_f, \epsilon_t)$} \\
        % \cline{2-19} 
        \cmidrule(rl){3-23}
         & & \multicolumn{3}{c}{$(0.3, 0.3, 0.3)$} 
           & \multicolumn{3}{c}{$(0.3, 0.3, 0.2)$}
           & \multicolumn{3}{c}{$(0.2, 0.3, 0.3)$} 
           & \multicolumn{3}{c}{$(0.2, 0.3, 0.2)$} 
           & \multicolumn{3}{c}{$(0.2, 0.3, 0.15)$} 
           & \multicolumn{3}{c}{$(0.2, 0.3, 0.1)$}
           & \multicolumn{3}{c}{$(0.2, 0.3, 0.4)$}\\
        % \cline{2-19} 
        \cmidrule(rl){3-5} \cmidrule(rl){6-8} \cmidrule(rl){9-11} \cmidrule(rl){12-14} \cmidrule(rl){15-17} \cmidrule(rl){18-20} \cmidrule(rl){21-23}
         & & $k=1$ & $k=2$ & $k=3$ & $k=1$ & $k=2$ & $k=3$ & $k=1$ & $k=2$ & $k=3$ 
           & $k=1$ & $k=2$ & $k=3$ & $k=1$ & $k=2$ & $k=3$ & $k=1$ & $k=2$ & $k=3$ 
           & $k=1$ & $k=2$ & $k=3$\\ 
        \midrule
        \multirow{4}{*}{20}
         & M1 & \bf{0.0212} & \bf{0.0276} &     0.0367  & \bf{0.0196} & \bf{0.0229} & 0.0273 & \bf{0.0212} & \bf{0.0276} & 0.0367 & \bf{0.0196} & \bf{0.0229} & 0.0273 & \bf{0.0189} & \bf{0.0211} & \bf{0.0238} & \bf{0.0183} & \bf{0.0195} & \bf{0.0210} & \bf{0.0232} & 0.0333 & 0.0483\\
         & M2 &     0.0513  &     0.0518  &     0.0523  &     0.0512  & 0.0515 & 0.0518 &     0.0513 & 0.0518 & 0.0523 & 0.0512 & 0.0515 & 0.0518 & 0.0512 & 0.0514 & 0.0516 & 0.0511 & 0.0512 & 0.0514 & 0.0515 & 0.0521 & 0.0528\\
         & M3 &     0.0251  & \bf{0.0279} & \bf{0.0343} &     0.0220  & 0.0236 & \bf{0.0264} &     0.0251 & \bf{0.0276} & \bf{0.0347} & 0.0216 & \bf{0.0232} & \bf{0.0259} & 0.0206 & 0.0216 & \bf{0.0236} & 0.0199 & 0.0204 & 0.0216 & 0.0286 & 0.0331 & \bf{0.0468}\\
         & M4 &     0.0260  & \bf{0.0278} &     0.0391  &     0.0270  & 0.0282 & 0.0333 &     0.0259  & \bf{0.0277} & 0.0374 & 0.0259 & 0.0273 & 0.0316 & 0.0262 & 0.0273 & 0.0298 & 0.0262 & 0.0271 & 0.0263 & 0.0261 & \bf{0.0283} & 0.0493\\
        \hline
        \multirow{4}{*}{50}
         & M1 & \bf{0.0272} &     0.0433  &     0.0662  & \bf{0.0231} & 0.0315 & 0.0426 & \bf{0.0272} & 0.0434 & 0.0662 & \bf{0.0231} & 0.0316 & 0.0427 & \bf{0.0214} & 0.0269 & 0.0338 & \bf{0.0198} & \bf{0.0230} & 0.0267 & \bf{0.0321} & 0.0580 & 0.0958\\
         & M2 &     0.0520  &     0.0532  &     0.0544  & 0.0517 & 0.0524 & 0.0532 & 0.0520 & 0.0532 & 0.0544 & 0.0517 & 0.0524 & 0.0532 & 0.0515 & 0.0521 & 0.0526 & 0.0513 & 0.0517 & 0.0521 & 0.0524 & 0.0540 & \bf{0.0557}\\
         & M3 &     0.0311  & \bf{0.0377} & \bf{0.0523} & 0.0258 & \bf{0.0292} & \bf{0.0381} & 0.0312 & \bf{0.0377} & \bf{0.0507} & 0.0253 & \bf{0.0291} & \bf{0.0358} & 0.0231 & \bf{0.0256} & \bf{0.0303} & 0.0214 & \bf{0.0229} & \bf{0.0256} & 0.0371 & 0.0474 & 0.0840\\
         & M4 &     0.0348  &     0.0388  &     0.0578  & 0.0338 & 0.0368 & 0.0526 & 0.0350 & 0.0391 & 0.0559 & 0.0333 & 0.0366 & 0.0447 & 0.0325 & 0.0353 & 0.0379 & 0.0311 & 0.0334 & 0.0302 & 0.0371 & \bf{0.0416} & 0.0821\\
        \hline
        \multirow{4}{*}{100}
         & M1 & \bf{0.0371} &     0.0689  &     0.1147  & \bf{0.0289} & 0.0454 & 0.0675 & \bf{0.0372} & 0.0690 & 0.1145 & \bf{0.0290} & 0.0456 & 0.0678 & \bf{0.0254} & 0.0362 & 0.0499 & \bf{0.0224} & 0.0286 & 0.0360 & \bf{0.0468} & 0.0984 & 0.1737\\
         & M2 &     0.0531  &     0.0555  & \bf{0.0579} & 0.0524 & 0.0540 & \bf{0.0555} & 0.0532 & 0.0555 & \bf{0.0579} & 0.0524 & 0.0540 & \bf{0.0555} & 0.0520 & 0.0532 & 0.0544 & 0.0517 & 0.0524 & 0.0532 & 0.0539 & \bf{0.0571} & \bf{0.0605}\\
         & M3 &     0.0390  & \bf{0.0496} &     0.0958  & 0.0311 & \bf{0.0379} & 0.0582 & 0.0395 & \bf{0.0495} & 0.0760 & 0.0308 & \bf{0.0382} & \bf{0.0553}& 0.0271 & \bf{0.0325} & \bf{0.0429} & 0.0238 & \bf{0.0271} & \bf{0.0327} & 0.0494 & 0.0612 & 0.1014\\
         & M4 &     0.0510  &     0.0613  &     0.0920  & 0.0449 & 0.0523 & 0.0579 & 0.0505 & 0.0643 & 0.0889 & 0.0438 & 0.0527 & 0.0600 & 0.0400 & 0.0455 & 0.0464 & 0.0346 & 0.0519 & 0.0360 & 0.0576 & 0.0689 & 0.1112\\
        \bottomrule
    \end{tabular}
\end{table}
\end{landscape}

\bibliographystyle{ACM-Reference-Format}
% \balance 
\bibliography{references}
\end{document}